\documentclass[10pt,twocolumn,letterpaper]{article}

\usepackage{iccv}
\usepackage{times}
\usepackage{epsfig}
\usepackage{graphicx}
\usepackage{amsmath}
\usepackage{amssymb}
\usepackage{subcaption}
\usepackage{multirow}

\usepackage[pagebackref=true,breaklinks=true,letterpaper=true,colorlinks,bookmarks=false]{hyperref}

\iccvfinalcopy 

\def\iccvPaperID{****} 

\ificcvfinal\fi
\begin{document}

\title{GANDiffFace: Controllable Generation of Synthetic Datasets\\for Face Recognition with Realistic Variations}

\author{Pietro Melzi$^1$
\and
Christian Rathgeb$^{2,3}$
\and
Ruben Tolosana$^1$
\and
Ruben Vera-Rodriguez$^1$
\and
Dominik Lawatsch$^2$
\and
Florian Domin$^2$
\and
Maxim Schaubert$^2$
\and
{\tt\small $^1$Biometrics and Data Pattern Analytics Laboratory, Universidad Autonoma de Madrid, Spain}
\and
{\tt\small $^2$secunet Security Networks AG, Essen, Germany}
\and
{\tt\small $^3$Hochschule Darmstadt, Germany}
}

\maketitle
\ificcvfinal\thispagestyle{empty}\fi

\begin{abstract}
    Face recognition systems have significantly advanced in recent years, driven by the availability of large-scale datasets. However, several issues have recently came up, including privacy concerns that have led to the discontinuation of well-established public datasets. Synthetic datasets have emerged as a solution, even though current synthesis methods present other drawbacks such as limited intra-class variations, lack of realism, and unfair representation of demographic groups.
    This study introduces GANDiffFace, a novel framework for the generation of synthetic datasets for face recognition that combines the power of Generative Adversarial Networks (GANs) and Diffusion models to overcome the limitations of existing synthetic datasets. In GANDiffFace, we first propose the use of GANs to synthesize highly realistic identities and meet target demographic distributions. Subsequently, we fine-tune Diffusion models with the images generated with GANs, synthesizing multiple images of the same identity with a variety of accessories, poses, expressions, and contexts.  
    We generate multiple synthetic datasets by changing GANDiffFace settings, and compare their mated and non-mated score distributions with the distributions provided by popular real-world datasets for face recognition, i.e. VGG2 and IJB-C. Our results show  the feasibility of the proposed GANDiffFace, in particular the use of Diffusion models to enhance the (limited) intra-class variations provided by GANs towards the level of real-world datasets. 
\end{abstract}

\section{Introduction}

In recent years, the development of face recognition technology has experienced a significant increase in the use of synthetic datasets. This trend has been facilitated by the proposal of numerous approaches for the generation of synthetic faces, resulting in an augmentation and diversification of the datasets for face recognition \cite{zhang2021applicability, kortylewski2018training}. 

Synthetic datasets provide several advantages compared to real-world datasets \cite{joshi2022synthetic}. Firstly, they offer a promising solution to some privacy concerns associated with real datasets, which are usually based on the collection of face images of individuals without their knowledge or consent from various online sources \cite{murgia2019s}.
Secondly, synthetic face generators provide potentially infinite data. This is of particular importance because established datasets have been dismissed due to privacy concerns \cite{Exposing_ai}, and regulatory frameworks such as the EU-GDPR require the informed consent of individuals prior to the collection and use of personal data \cite{voigt2017eu}. Finally, if the synthesis process is controllable, datasets with desired demographic characteristics (and labels for free) can be easily obtained, unlike real-world datasets that unequally represent diverse demographic groups \cite{morales2020sensitivenets}, among other aspects. 

\begin{figure}[t]
\begin{center}
    \includegraphics[width=1\linewidth]{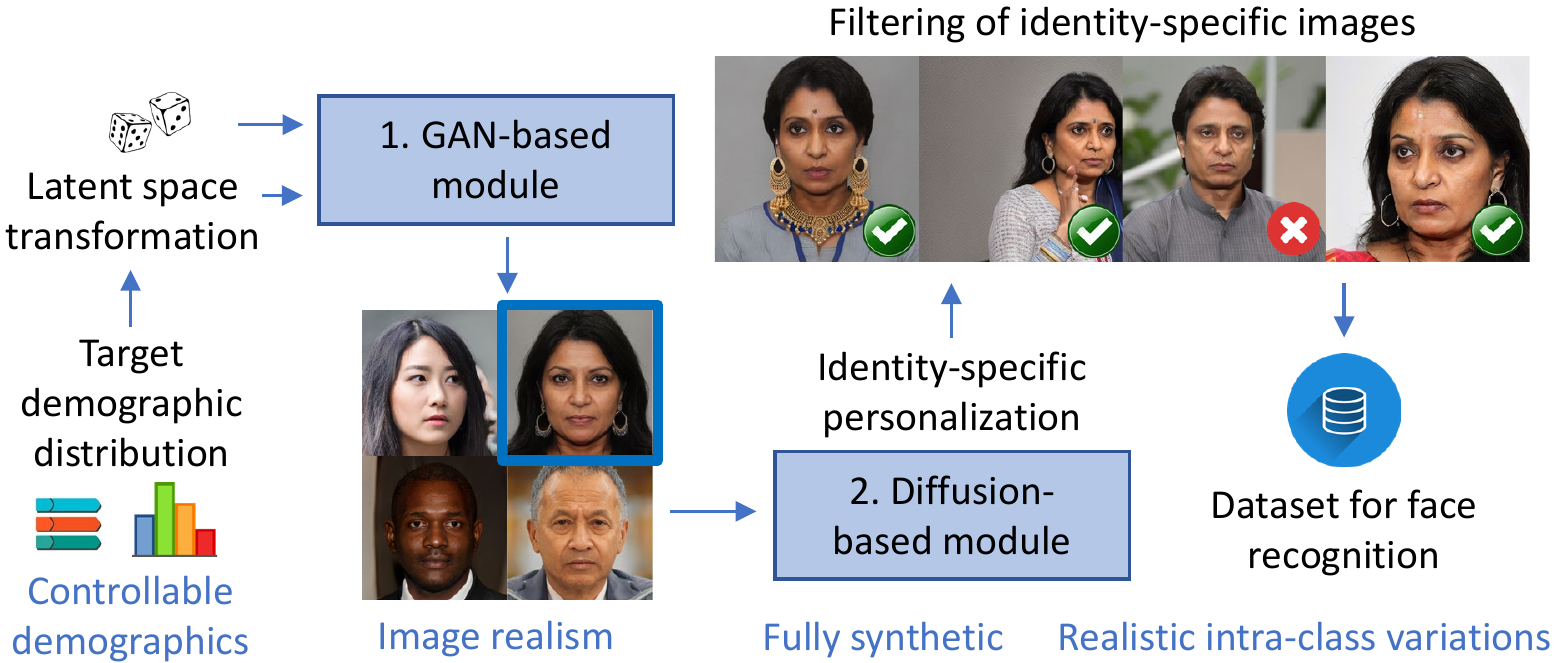}
\end{center}\vspace{-0.4cm}
\caption{Overview of our GANDiffFace framework based on the combination of GAN and Diffusion models. GANDiffFace creates synthetic datasets for face recognition with the properties listed in blue. From each identity synthesized with the GAN-based module, a personalized Diffusion-based module generates images with realistic intra-class variations that, once filtered, will compose the final dataset.}\vspace{-0.4cm}
\label{fig:intro}
\end{figure}

Among generative models, Generative Adversarial Networks (GANs) have been widely used to synthesize face images due to their ability to learn complex distributions and generate high-quality images of human faces \cite{kammoun2022generative, saxena2021generative}, especially the recent version of StyleGAN3 \cite{karras2021alias, alaluf2023third}. However, GANs generate images based on the patterns learned from the training data, with limited control over the generated features, and possible biases towards certain demographic groups over-represented during training \cite{maluleke2022studying}. To tackle this issue, some methods have been proposed in the literature to modify attributes of synthetic face images, \emph{i.e.} pose, illumination, and demographics. Target attributes can be injected into the generative component of GANs \cite{chu2020sscgan, georgopoulos2021mitigating}, or alternatively the latent structure of GANs, \emph{i.e.} their internal representation of face images, can be properly manipulated to meet the target attributes \cite{shen2020interpreting, yang2021l2m}. However, GAN-generated images have been found to exhibit insufficient variability between the images of the same individual (\emph{i.e.} intra-class variation), in comparison to real-world datasets \cite{colbois2021use}. This affects the performance of face recognition models trained with synthetic data and evaluated with real data, as observed in \cite{qiu2021synface}.

More recently, Diffusion models have gained popularity and outperformed GAN models in multiple tasks, including image synthesis \cite{dhariwal2021diffusion}. A Diffusion model consists of a Markov chain that gradually adds random noise to data and learns to reverse it, to generate the desired output from noise \cite{ho2020denoising}. Diffusion models can shape their outputs according to text or images \cite{von-platen-etal-2022-diffusers}, and generate a wider variety of images compared to GAN models \cite{kingma2021variational}. However, unlike GANs, Diffusion models do not learn explicit latent representations of face images, making their demographic attributes and intra-class variations less controllable \cite{dhariwal2021diffusion}. 

In this study, we propose a novel framework called \emph{GANDiffFace} to generate synthetic datasets for face recognition, by combining the advantages of both GAN and Diffusion models (Figure \ref{fig:intro}). 
We use StyleGAN3 to generate synthetic identities, and create six different images for each identity by manipulating their pose, expression, and illumination attributes in the latent space. For attribute manipulation, we follow the approach (detailed in Section \ref{sec:3.1}) proposed by a previous work that investigates the use of automatically generated synthetic datasets for benchmarking face recognition systems \cite{colbois2021use}. We observe that such synthetic datasets are not suitable for other tasks, \emph{e.g.} the training of face recognition systems, because of their limited intra-class variations. 
Hence, we propose the use of DreamBooth, a recent framework for the ``personalization'' of Diffusion models \cite{ruiz2023dreambooth}, to generate more realistic intra-class variations. Given as input the six images previously generated for a specific subject, DreamBooth fine-tunes a pretrained text-to-image Diffusion model to bind a unique identifier with that subject. The unique identifier allows to synthesize fully-novel photorealistic images of the subject contextualized in different scenes, poses, views, and lighting conditions, by leveraging the semantic prior embedded in the model \cite{ruiz2023dreambooth}. 

The main contributions of the study are:
\begin{itemize}
    \item Proposal of GANDiffFace, a novel framework for the generation of synthetic datasets for face recognition. GANDiffFace generates photorealistic images of synthetic identities with enhanced intra-class variations. Additionally, specific demographic distributions can be obtained by manipulating the latent space of StyleGAN3 during identity generation.
    \item Two different datasets with the same synthetic identities are generated at different steps of GANDiffFace: \emph{i)} with the GAN-based module alone, and \emph{ii)} with the combination of GAN-based and Diffusion-based modules. We provide a direct comparison (based on the same identities) between the two synthetic datasets, and further compare them to real-world datasets.
    \item We make available the synthetic dataset generated with GANDiffFace, characterized by easily controllable and realistic intra-class variations. Our dataset represents equally balanced demographic groups, defined in terms of race, age, and gender, and contains labels of several face attributes. Hence, it enables the training/testing of multiple facial analysis applications.
\end{itemize}

The remainder of this work is organized as follows: in Section \ref{sec:2} we describe related works that use synthetic datasets for face recognition. In Section \ref{sec:3} we describe the modules of our proposed GANDiffFace framework. In Section \ref{sec:4} we provide an evaluation on our synthetic datasets, and in Section \ref{sec:5} we discuss limitations and future works, drawing the conclusions of this work.

\section{Related works} \label{sec:2}
Numerous technologies have been proposed to generate synthetic datasets for face recognition. The applicability of synthetic datasets to face recognition has been investigated in \cite{zhang2021applicability}, to compensate for the lack of publicly available large-scale test datasets, and in \cite{boutros2023synthetic}, to provide a taxonomy and further discussion.
In Table \ref{tab:related}, we compare the most relevant synthetic datasets for face recognition proposed in the literature. 

\begin{table*}
\begin{center}
\begin{small}
\begin{tabular}{|c|c|c|c|c|c|}
\hline
\textbf{Method}   & \textbf{Category}                                                     & \textbf{Realism} & \textbf{\begin{tabular}[c]{@{}c@{}}Controllable\\demographics\end{tabular}} & \textbf{\begin{tabular}[c]{@{}c@{}}Intra-class\\ variations\end{tabular}} & \textbf{\begin{tabular}[c]{@{}c@{}}Fully\\synthetic\end{tabular}} \\ \hline \hline
Latent space \cite{colbois2021use}      & GAN                                                             &      high            &                           low                                               &                  low                  & yes                                       \\ 
HDA-SynChildFaces \cite{falkenberg2023child} & GAN                                                             &     high             &                                  high                                        &                     low               & yes                                       \\ 
SYNFace \cite{qiu2021synface}          & GAN                                                             &     high             &                                       low                                   &                       low                  & no                                  \\ 
SFace \cite{boutros2022sface}          & GAN                                                             &     high             &                                       low                                   &                       low                  & yes                                  \\ 
DigiFace-1M \cite{bae2023digiface}       & 3D model                                                              &        low          &                             medium                                             &                high                      & yes                                     \\ 
DCFace \cite{kim2023dcface}            & Diffusion                                                       &        medium          &                                       low                                   &                           high              & no                                  \\ \hline
\textbf{GANDiffFace (ours)}             & \textbf{GAN + Diffusion} &        \textbf{high}          &                         \textbf{high}                                                 &                            \textbf{high}                & \textbf{yes}                               \\ \hline
\end{tabular}
\end{small}
\end{center}\vspace{-0.4cm}
\caption{Overview of the synthetic datasets for face recognition applications proposed in the literature.}\vspace{-0.4cm}
\label{tab:related}
\end{table*}

StyleGAN2 is used to generate synthetic identities in \cite{colbois2021use}. With the property of \emph{linear separability} of StyleGAN2's latent space, multiple images of the original identities are generated while changing three attributes, \emph{i.e.} illumination, pose, and expression. Linear separability allows to find a hyperplane in the latent space that separates populations of latent vectors according to different values for a specific attribute. The normal vector to this hyperplane represents the direction along which latent vectors, \emph{i.e.} the representations of synthetic images in the latent space, can be moved to modify the specific attribute. 
The approach proposed in \cite{colbois2021use} presents some limitations addressed by our GANDiffFace, namely the demographic bias inherited from StyleGAN2, and the limited intra-class variations generated. 

In an analogous way, the \emph{linear separability} of StyleGAN3's latent space is exploited to generate a large-scale synthetic dataset of children's faces, named \emph{HDA-SynChildFaces} \cite{falkenberg2023child}. Compared to the previous work, in \emph{HDA-SynChildFaces} the latent space is manipulated during identity generation to balance the race distribution of the dataset. The work reveals that children consistently perform worse than adults in various face recognition systems. 

\emph{SYNFace} proposes the use of DiscoFaceGAN for the synthesis of face images, a disentangled learning scheme that enables precise control of targeted face properties such as identity, pose, expression, and illumination \cite{qiu2021synface}. DiscoFaceGAN generates realistic face images by sampling random noise from multiple normal distributions, each one independently controlling a different face attribute. \emph{SYNFace} identifies in poor intra-class variations the reason of the performance gap existing between face recognition systems trained with synthetic and real datasets. To mitigate it, the intermediate states of two synthetic identities mixed together are considered as novel identities. \emph{SYNFace} generates mostly frontal-view images, the identity preservation or variation of mixed identities is not evaluated, and a further mix with real images is required to bridge the gap between synthetic and real world data.

In \emph{SFace} a privacy-friendly synthetically generated face dataset is proposed, based on the training of StyleGAN2-ADA with real datasets, and the setting of identity labels as class labels to create synthetic data \cite{boutros2022sface}. Hence, a 1:1 correspondence can be observed between real and synthetic identities, with the consequent sharing of face attributes (but not identity). \emph{SFace} provides an unrealistic mated score distribution, shifted towards the non-mated distribution, and unlike other GAN-based methods maintains a tight correspondence between the synthetic identities and the real ones used during training. 

\emph{DigiFace-1M}, a large-scale synthetic dataset obtained by rendering digital faces with a computer graphics pipeline, is proposed for face recognition in \cite{bae2023digiface}. Each identity of \emph{DigiFace-1M} is defined as the unique combination of facial geometry, texture, eye color, and hair style, while other parameters (\emph{i.e.} pose, expression, environment, and camera distance) are varied to render multiple images. With aggressive data augmentation, this work significantly reduces the synthetic-to-real domain gap, establishing the new state-of-the-art performance for face recognition models trained on synthetic data. Furthermore, this method does not rely on real data for training the generative model, differently from GAN models.
However, we observe some limitations in \emph{DigiFace-1M}: the textures of the synthetic images appear unrealistic, and the demographic distribution of the synthetic dataset is not analyzed.

\begin{figure*}[t]
\begin{center}
\begin{subfigure}{\textwidth}
    \centering
    \includegraphics[width=1\linewidth]{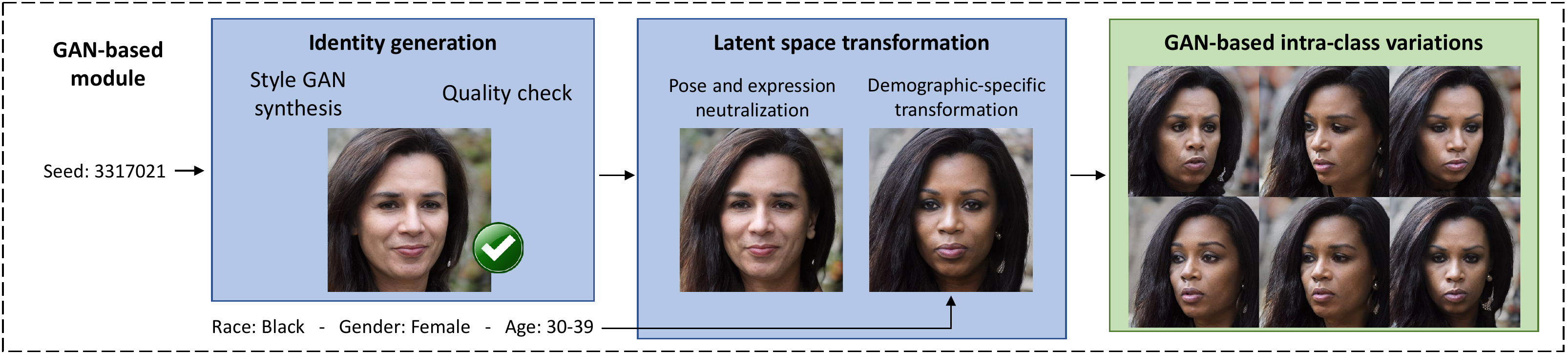}
    \label{fig:framework_a}\vspace{-0.4cm}
    \caption{The GAN-based module of GANDiffFace. Given a seed and target demographic attributes, multiple images of the same identity are provided with GAN-based (limited) intra-class variations.}
\end{subfigure}
\begin{subfigure}{\textwidth}
    \centering
    \includegraphics[width=1\linewidth]{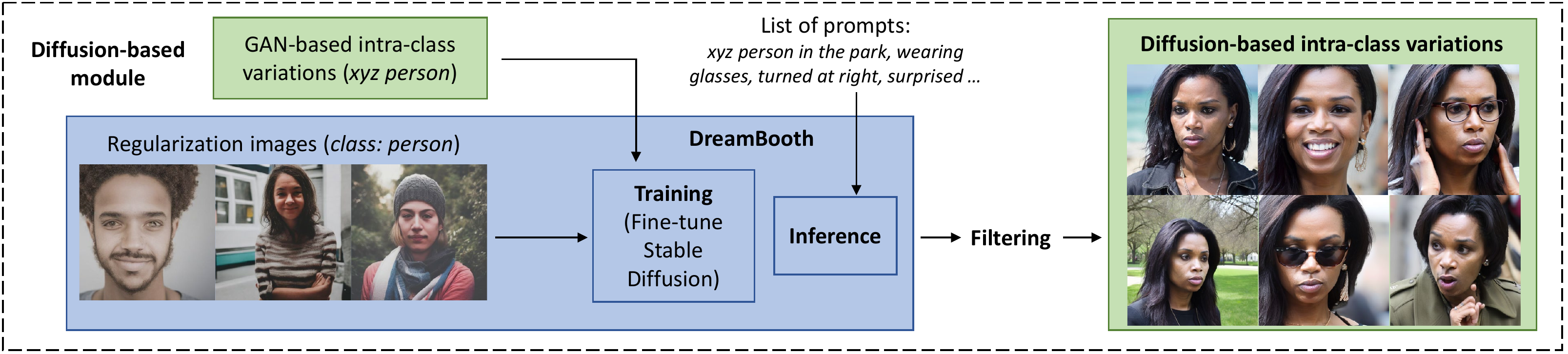}
    \caption{The Diffusion-based module of GANDiffFace. Given a few images with GAN-based (limited) intra-class variations (a) of the synthetic identity \emph{xyz}, a set of images of the class \emph{person} for regularization (generated by Diffusion model), and a list of prompts, multiple images of the same identity are provided with augmented Diffusion-based intra-class variations.}\vspace{-0.4cm}
    \label{fig:framework_b}
\end{subfigure}
\end{center}
\caption{Graphical representation of the GAN-based (a) and Diffusion-based (b) modules of the GANDiffFace framework.}\vspace{-0.4cm}
\label{fig:framework}
\end{figure*}

More recently, a Diffusion model called \emph{DCFace} has been proposed for synthetic face recognition \cite{kim2023dcface}. \emph{DCFace} is composed of: \emph{i)} a sampling stage for the generation of synthetic identities, and \emph{ii)} a mixing stage for the generation of face images whose identity comes from the sampling stage and the style is selected from a ``style bank'' of images. Both components are based on Diffusion models, showing considerable ability to generate unique and diverse identities. Compared to \emph{SYNFace} and \emph{DigiFace-1M}, \emph{DCFace} claims to provide better intra-class variations, but relies on real face images for the ``style bank''. While synthetic data could in principle also be used for the ``style bank'', this may reduce intra-class variations in the generated dataset. We raise criticism about the use of real data, as newly generated synthetic images contain sharp details from the real images used as style reference and the method is not fully synthetic. Furthermore, with \emph{DCFace} specific face attributes cannot be manipulated either during the sampling or mixing stages. 

As of today, synthetic datasets based on Diffusion models are promising, but still in a primitive stage. Our proposed GANDiffFace framework combines the advantages of GAN models, \emph{i.e.} generation of highly realistic faces and control of the latent space, with enhanced intra-class variations achieved by recent Diffusion models. 

\section{Proposed method} \label{sec:3}

The graphical representation of our proposed GANDiffFace framework is provided in Figure \ref{fig:framework}. GANDiffFace consists of two modules: the first one dedicated to the synthesis of identities, based on StyleGAN3 \cite{karras2021alias} and transformation in its latent space (Section \ref{sec:3.1}), and the second one responsible for the creation of realistic intra-class variations, based on DreamBooth \cite{ruiz2023dreambooth} (Section \ref{sec:3.2}). 

\subsection{GAN-based module}
\label{sec:3.1}

\begin{figure}[t]
\begin{center}
\begin{subfigure}{0.16\textwidth}
    \centering
    \includegraphics[width=1\linewidth]{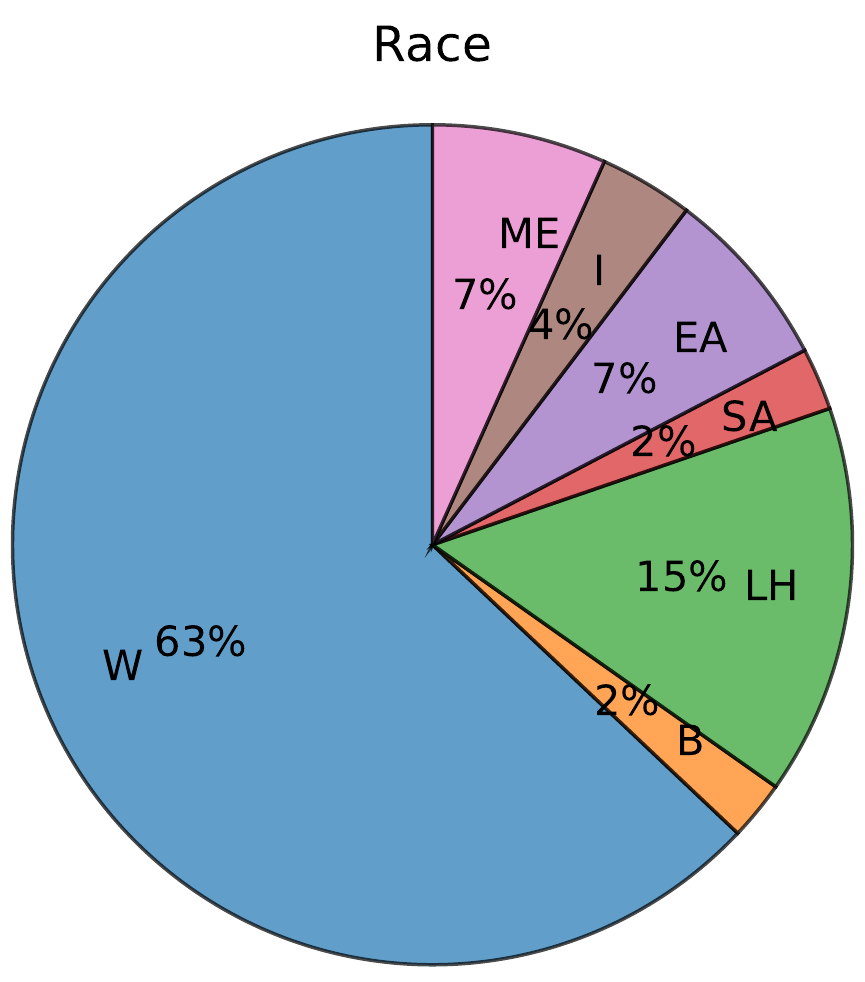}
\end{subfigure}%
\begin{subfigure}{0.16\textwidth}
    \centering
    \includegraphics[width=1\linewidth]{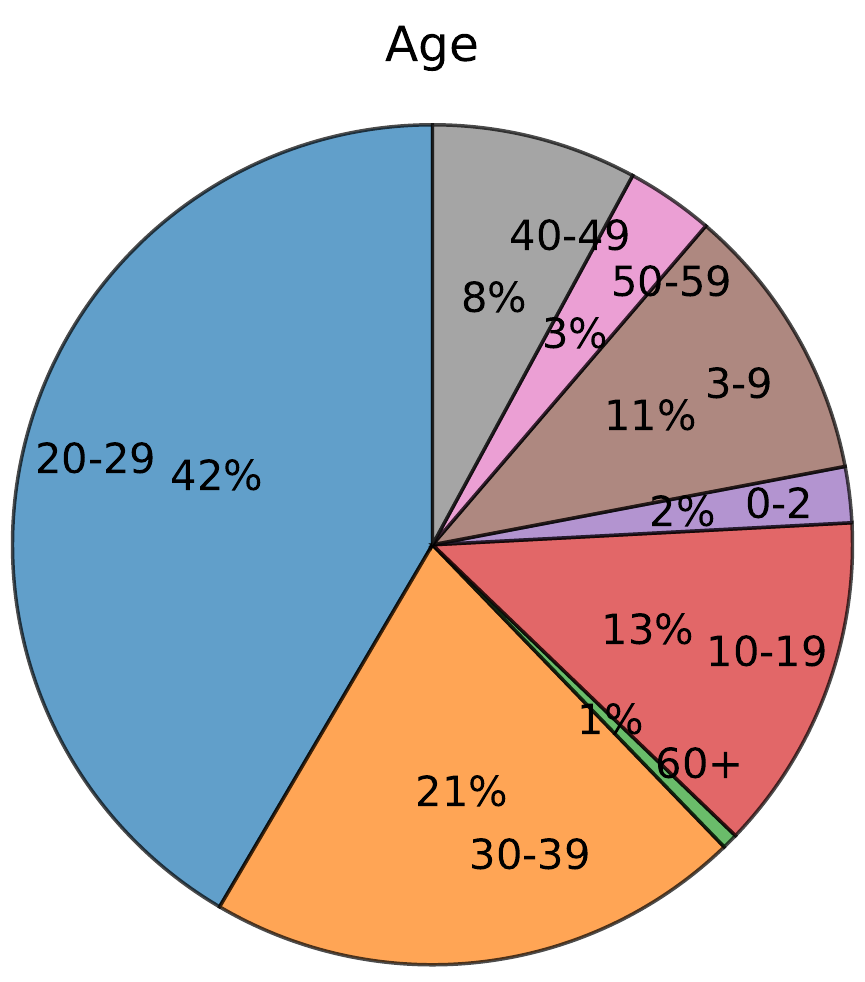}
\end{subfigure}%
\begin{subfigure}{0.16\textwidth}
    \centering
    \includegraphics[width=1\linewidth]{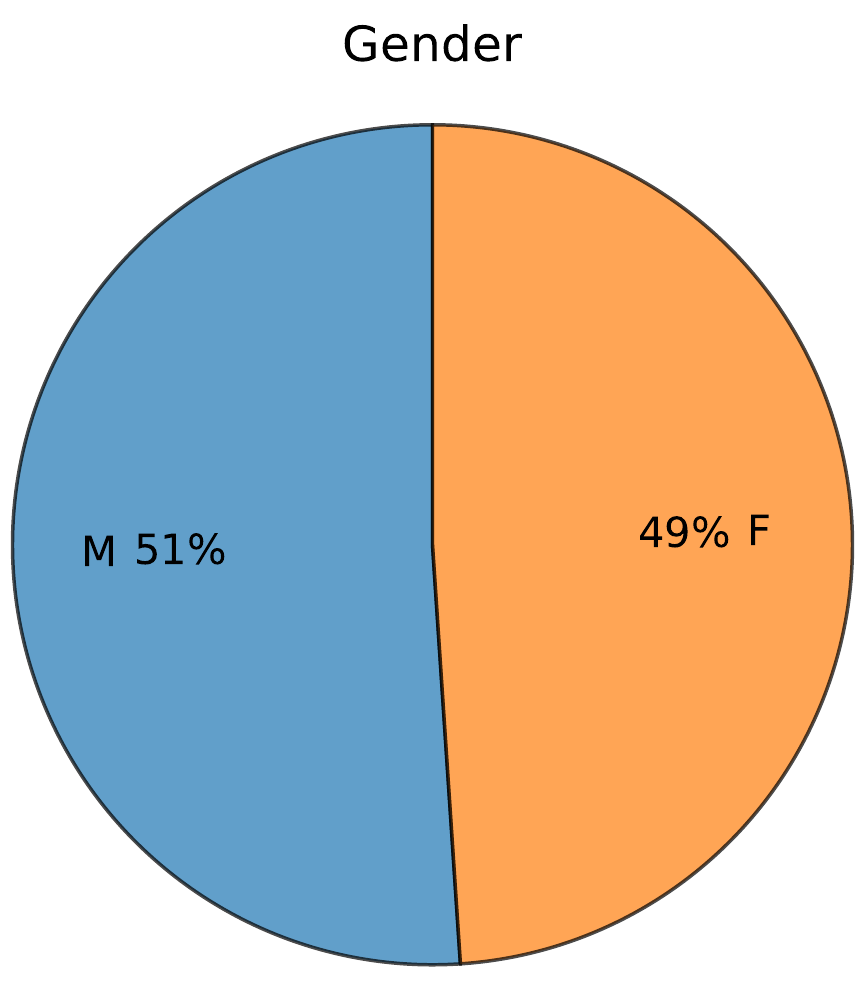}
\end{subfigure}
\end{center}\vspace{-0.4cm}
\caption{Race (W=White, B=Black, LH=Latino\_Hispanic, SA=Southeast Asian, EA=East Asian, ME=Middle Eastern, I=Indian), age, and gender (M=male, F=Female) distributions in the initial random set of 256,000 identities generated with StyleGAN3 \cite{karras2021alias} and labelled with FairFace \cite{karkkainen2021fairface}.}\vspace{-0.4cm}
\label{fig:demographic}
\end{figure}

\paragraph{Identity generation.}
We first generate an initial random set of 256,000 synthetic images with StyleGAN3 (pre-trained with FFHQ dataset \cite{karras2019style}), and label them with FairFace, a classifier of demographic attributes (\emph{i.e.} race, gender, and age) \cite{karkkainen2021fairface}. The distributions of demographic attributes obtained in the random set are reported in Figure \ref{fig:demographic}, highlighting the bias present in StyleGAN3. We remove from the initial set images with poor quality as well as those belonging to young subjects. For quality assessment we use MagFace with backbone iResNet100, a state-of-the-art system that learns feature embeddings whose magnitudes represent face sample quality \cite{meng2021magface}. We eliminate the 10\% of images with the lowest magnitude, that usually contain artifacts, sunglasses, or belong to children. Then, we also eliminate images of people in the age intervals 0-2, 3-9, and 10-19, as we focus only on adult identities.

\paragraph{Face attribute representation.}
A framework to interpret the disentangled face representation learned by StyleGAN and study the properties of the facial semantics encoded in its latent space was initially proposed in \cite{shen2020interpreting}. The framework is based on the training of linear Support Vector Machines (SVMs) in the latent space to separate two distinct populations of latent vectors according to a binary target attribute. The normal vector to the resulting hyperplane boundary of the trained SVM represents the direction to follow in the latent space to edit the target attribute of face images. This approach has proved successful even in the latent space of StyleGAN2 \cite{colbois2021use} and StyleGAN3 \cite{falkenberg2023child}.

In this work, we label our synthetic dataset according to pose (yaw and pitch) with 3DDFA\_V2 \cite{guo2020towards, 3ddfa_cleardusk}, expression (neutral, happy, sad, surprise, disgust, anger, contempt) with DMUE \cite{she2021dive}, and illumination by comparing the pixel intensity of the right and left half of face images. We also consider the labels provided by FairFace for gender, age, and race. For each attribute of interest, we represent two populations with an equal number of latent vectors, selected at the two extremes of the score distribution of the target attribute. We train each SVM with a maximum number of 100,000 latent vectors, depending for each attribute on the amount of data available to represent populations. In case of categorical attributes, \emph{i.e.} expression and race, numerical values are provided respectively by DMUE and FairFace for all the possible categorical attributes. Hence, we train multiple one-vs-one SVMs to separate each expression from the neutral one, and multiple one-vs-all SVMs for each different race. In Table \ref{tab:2} we report all the boundaries calculated in this work, providing additional information about the training of each SVM. High validation accuracy demonstrates the goodness of our boundaries, except for \emph{illumination} that turns out to be unreliable. The entire training of boundaries is carried out exclusively with synthetic data. 

\begin{table}
\begin{center}
\begin{small}
\setlength{\tabcolsep}{4pt}
\begin{tabular}{|c|c|c|c|}
\hline
\textbf{Attribute} & \textbf{\begin{tabular}[c]{@{}c@{}}Number\\of images\end{tabular}} & \textbf{\begin{tabular}[c]{@{}c@{}}Validation\\ accuracy\end{tabular}} & \textbf{\begin{tabular}[c]{@{}c@{}}Average\\ distance\end{tabular}} \\ \hline\hline
Pose: Yaw               & 100,000                                                                & 100\%                                                                  & 1.39                                                                \\ 
Pose: Pitch             & 100,000                                                                & 99\%                                                                   & 0.98                                                                \\ \hline
Expression: Happy             & 13,390                                                                 & 100\%                                                                  & 1.11                                                                \\ 
Expression: Contempt          & 11,014                                                                 & 92\%                                                                   & 0.46                                                                \\ 
Expression: Surprise          & 8,328                                                                  & 89\%                                                                   & 0.53                                                                \\ 
Expression: Disgust           & 4,436                                                                  & 95\%                                                                   & 0.84                                                                \\ 
Expression: Sad               & 2,606                                                                  & 85\%                                                                   & 0.45                                                                \\ 
Expression: Anger             & 2,440                                                                  & 91\%                                                                   & 0.74                                                                \\ \hline
Illumination             & 15,000                                                                 & 72\%                                                                   & 0.18                                                                \\ \hline
Gender            & 100,000                                                                & 100\%                                                                  & 1.33                                                                \\ \hline
Age               & 37,736                                                                 & 96\%                                                                   & 0.85                                                                \\ \hline
Race: White             & 64,846                                                                 & 100\%                                                                  & 1.00                                                                \\ 
Race: Latino-Hispanic   & 33,762                                                                 & 98\%                                                                   & 0.92                                                                \\ 
Race: East Asian        & 12,964                                                                 & 100\%                                                                  & 1.12                                                                \\ 
Race: Middle Eastern    & 13,112                                                                 & 92\%                                                                   & 0.57                                                                \\ 
Race: Indian            & 8,244                                                                  & 100\%                                                                  & 1.59                                                                \\ 
Race: Southeast Asian   & 5,180                                                                  & 100\%                                                                  & 1.43                                                                \\ 
Race: Black             & 5,356                                                                  & 100\%                                                                  & 1.92                                                                \\ \hline
\end{tabular}
\end{small}
\end{center}\vspace{-0.4cm}
\caption{List of boundaries calculated in this work, with information about each SVM training. Average distance is the distance of latent vectors from the hyperplane boundary.}\vspace{-0.2cm}
\label{tab:2}
\end{table}

\paragraph{Latent space transformation.}
The approach used to modify face attributes by applying transformations in the latent space has been described in detail in previous works \cite{colbois2021use, falkenberg2023child}. For clarity, here we only summarize its key points. We can transform a latent vector $w$, that represents a face image in the latent space of StyleGAN3, to modify its attribute $a$ according to the following operation:

\begin{figure}[t]
\begin{center}
\includegraphics[width=1\linewidth]{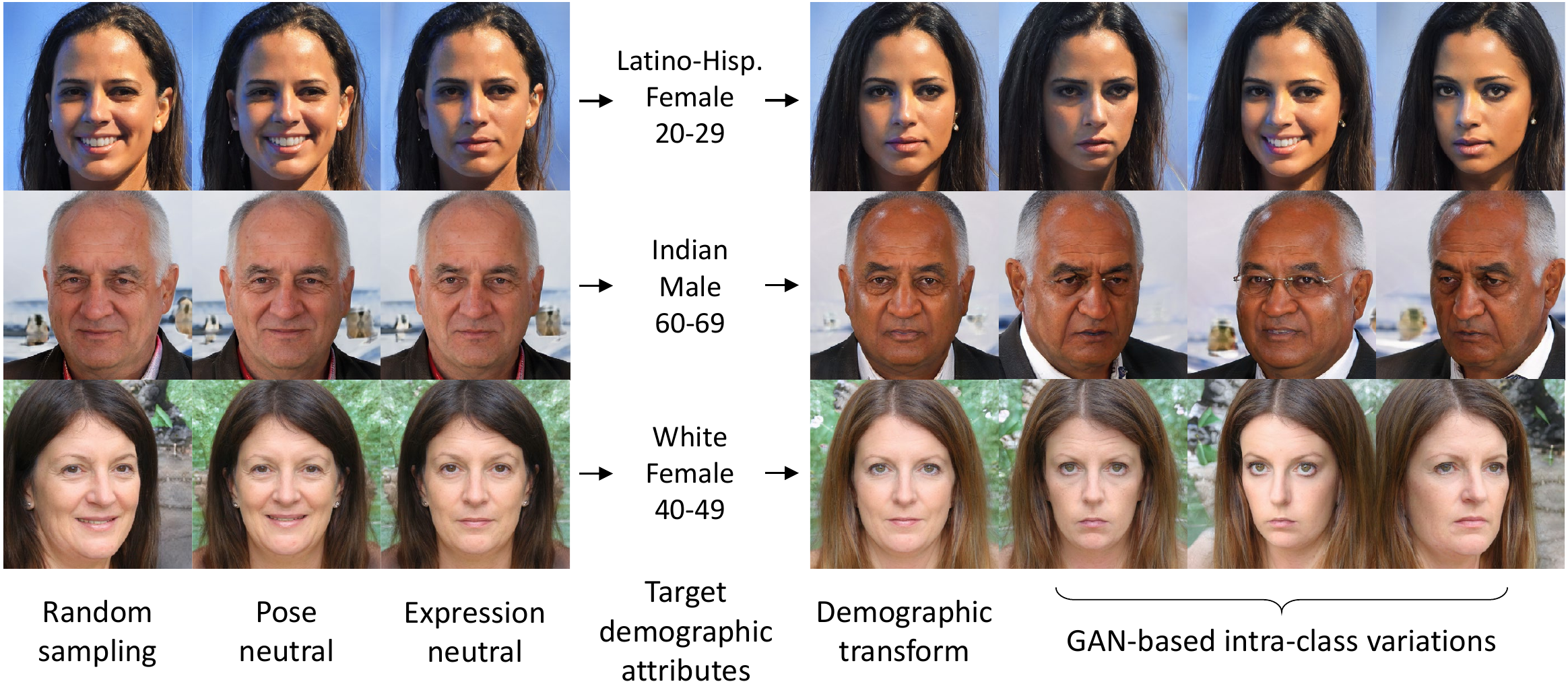}
\end{center}\vspace{-0.4cm}
\caption{Sequence of transformations to generate identities with target demographic attributes and GAN-based intra-class variations. Initial identities with demographic attributes similar to the target ones have been selected.}\vspace{-0.4cm}
\label{fig:sequences}
\end{figure}

\begin{equation}
    w_a = w + \alpha \cdot n_a, 
\end{equation}
where $n_a$ is the normal vector to the hyperplane that separates populations according to the attribute $a$, $\alpha$ is the degree of the transformation, and $w_a$ is the resulting latent vector, in which the attribute $a$ results modified according to the direction of the transformation. To neutralize a latent vector $w$ with respect to the attribute $a$, it is possible to project $w$ onto the hyperplane boundary of attribute $a$, as follows:

\begin{equation}
    w_{n_a} = w - (w^T n_a) \cdot n_a, 
\end{equation}
where $w_{n_a}$ is the resulting neutralized latent vector, in which the attribute $a$ results in a neutral condition. By combining the operations of transformation and neutralization to modify the demographic attributes of face images (\emph{i.e.} race, age, and gender), an arbitrary large number of identities can be generated to represent target demographic groups. In the following we describe the sequence of operations required to generate identities with target demographic attributes, and provide (limited) intra-class variations with a GAN-based approach. In Figure \ref{fig:sequences}, we also provide graphical examples of these operations for random identities.

\begin{enumerate}
    \item \emph{Pose neutralization:} the pose of the random identities generated with StyleGAN3 is neutralized, by projecting their latent vectors on the hyperplane boundaries relative to yaw and pitch.
    \item \emph{Expression neutralization:} the expression of the random identities is neutralized, by projecting their latent vector on the hyperplane boundary relative to the current expression of each identity, and subsequently moving the resulting latent vectors in the direction of neutral expression (opposite direction with respect to current expression).
    \item \emph{Demographic-specific transformation:} the latent vectors (neutralized according to pose and expression) are modified by applying transformations in the direction of the boundaries of interest. The pre-selection of random identities with demographic attributes close to the target ones may help to prevent transformations from estimating latent vectors outside of the StyleGAN3 distribution of faces \cite{falkenberg2023child}. We consider $70$ different demographic groups, obtained by combining the seven races, five adult age intervals, and two genders reported in Figure \ref{fig:demographic}. In total, we generate at this step $700$ different identities ($10$ identities for each of the $70$ demographic groups).
    \item \emph{GAN-based intra-class variations:} the latent vectors of demographic-specific identities can be further modified according to the boundaries of pose, expression, and illumination, to generate (limited) intra-class variations for each synthetic identity. 
\end{enumerate}
 
We observe that kinship ties, multiethnic unions, and population aging can be simulated by applying different demographic transformations to the same original identity.

\subsection{Diffusion-based module}
\label{sec:3.2}

Text-to-image models enable high-quality and diverse synthesis of images based on text prompts. They rely on their strong semantic prior, learned from a large collection of image-caption pairs, to bind a word with various images in different poses and contexts \cite{ramesh2022hierarchical, saharia2022photorealistic}. However, these models lack the ability to preserve the identity of a subject consistently across synthesized images. To overcome this issue we consider Dreambooth, a novel framework that fine-tunes text-to-image models (in this case Stable Diffusion \cite{rombach2022high}) to bind new words with specific subjects, and synthesize novel renditions of subjects in different contexts while maintaining their distinctive features \cite{ruiz2023dreambooth}. 

\paragraph{Training.}

We use the images generated by the GAN-based module of GANDiffFace to fine-tune Stable Diffusion, a state-of-the-art Diffusion text-to-image model \cite{rombach2022high}. We apply Dreambooth to bind a unique token (we use \emph{xyz}) with a specific synthetic identity, and implant it into the output domain of Stable Diffusion. To refer to the identity, we use text prompts containing the token \emph{xyz} followed by the class name of the identity, in our case \emph{person}. Hence, the minimum text prompt to refer to the identity is: ``xyz person''. The class name (\emph{i.e.} person) enables the model to use its prior knowledge of the class, and an additional class-specific prior preservation loss helps to prevent the model to associate the class with the specific identity. These components serve as regularization, as they alleviate overfitting and encourage diversity in the resulting images \cite{ruiz2023dreambooth}.

Previous studies highlighted the importance of parameter settings to fine-tune the Stable Diffusion model, especially in case of the \emph{person} class \cite{hugging}. We fine-tune Stable Diffusion with 6 input images for each synthetic identity, 200 images of the class \emph{person} for regularization (generated by Stable Diffusion itself), and for 1,000 epochs, also allowing the fine-tuning of the text encoder. Given the high number of identities in our dataset and possible interferences between tokens in the vocabulary, we fine-tune a specific Stable Diffusion model for each synthetic identity. 

\paragraph{Inference.}

Once fine-tuned with DreamBooth, the Stable Diffusion model can generate images of the specific synthetic identity in multiple contexts, according to the provided prompts. To generate synthetic images with realistic intra-class variations, we evaluate different categories of prompts: accessorization, advanced poses, advanced expressions, and recontextualization.
Examples of prompts are: ``xyz person wearing scarf'', ``close photo of xyz person at the beach'', ``skeptical xyz person'', and ``full body xyz person with accurate details of face in an indoor place''.

We observe that Stable Diffusion allows to specify negative prompts, to prevent the generation of undesired images. Given the large variety of datasets used to train Stable Diffusion, for the entire inference process we specify the following negative prompt: ``photo with the style of painting, comics, drawing, or containing text''. 

\paragraph{Filtering.}
The quality of the images generated with text-to-image Diffusion models highly depends on the correct specification of text prompts \cite{xie2023prompt}. In our inference phase, we consider some prompts that may work well for most but not all the identities, to enhance the intra-class variations resulting in our dataset. For this reason, an important component of our GANDiffFace framework is the filtering of the generated images, which is carried out in three stages:

\begin{figure}[t]
\begin{center}
    \includegraphics[width=0.9\linewidth]{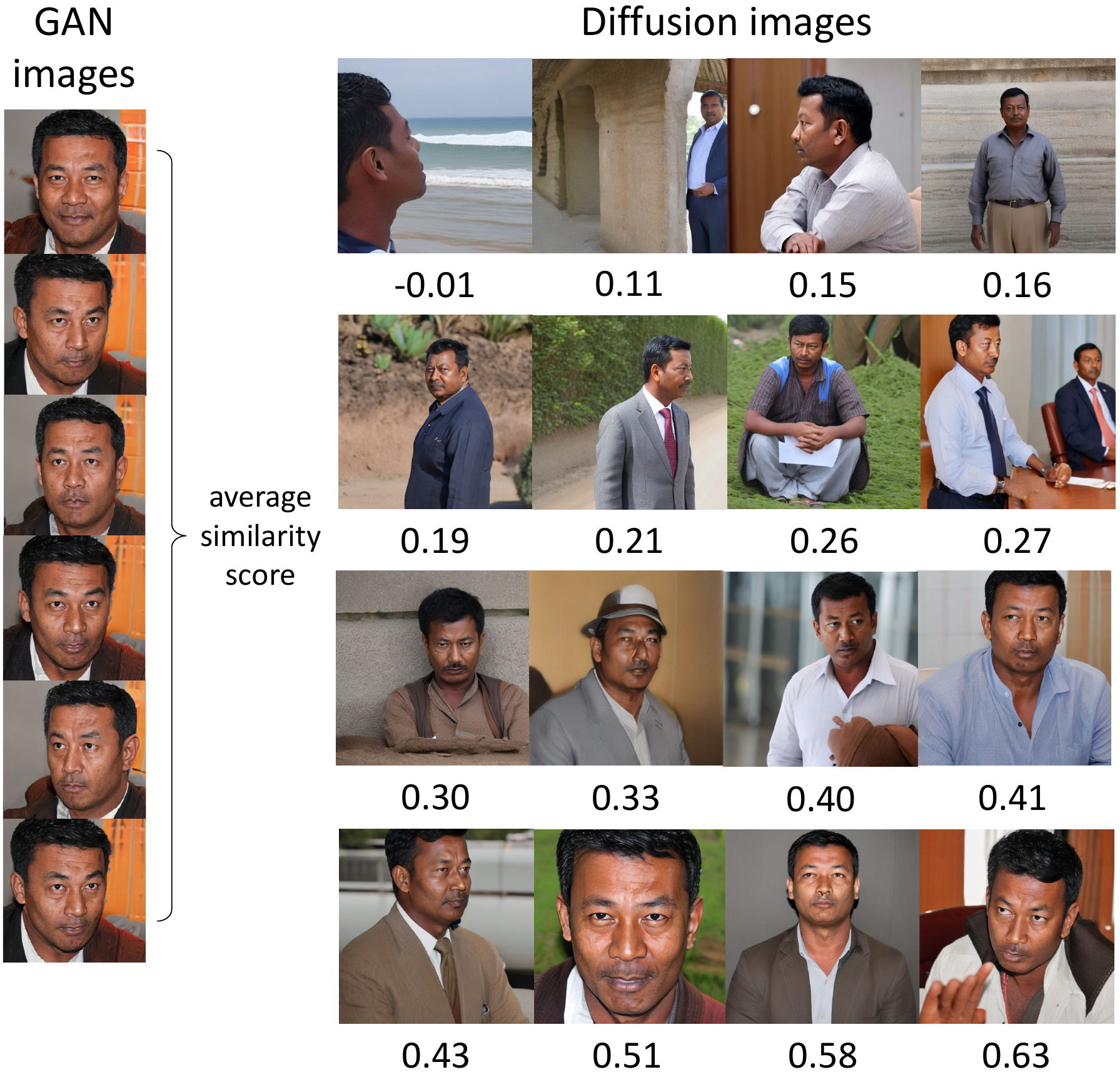}
\end{center}\vspace{-0.4cm}
\caption{Identity preservation scores, obtained for each image on the right by averaging their similarity scores with the GAN images on the left. Threshold $t_{ip}$ regulates the intra-class variations provided by GANDiffFace.}
\label{fig:identity_preservation}\vspace{-0.4cm}
\end{figure}

\begin{enumerate}
    \item \emph{Face detection:} we filter out images in which the face detector SCRFD-10G \cite{guo2021sample} detects no face.
    \item \emph{Identity preservation:} we extract ArcFace embeddings (backbone iResNet100) \cite{deng2019arcface} from synthetic images generated with both the GAN-based module only and the entire GANDiffFace framework. Then, for each synthetic image generated with  GANDiffFace, we calculate the average of its cosine similarity with the 6 GAN images that represent the same identity (previously used to fine-tune the Stable Diffusion model). We filter out images if the similarity score is below a threshold $t_{ip}=0.3$. 
    We note that similarity scores are computed between images of different domains, allowing the removal of outliers images generated with the Diffusion-based module, and no comparison is carried out between images of the same domain. In Figure \ref{fig:identity_preservation}, we include examples of the average similarity scores between GAN and Diffusion images for a random identity.
    \item \emph{Gender preservation:} we label the remaining images by gender with FairFace, and filter out images with a gender different from the corresponding GAN images. 
\end{enumerate}

\section{Evaluation} \label{sec:4}

This section analyzes the similarity scores obtained with four versions of our synthetic dataset, in order to provide a comparison with the score distributions of existing synthetic and real-world datasets used for face recognition. 

\subsection{Synthetic datasets}
We generate four datasets selecting different settings of our GANDiffFace framework, all of them containing the same $700$ synthetic identities. 
We only use the GAN-based module of GANDiffFace to generate a synthetic dataset provided with GAN-based (limited) intra-class variations. We use the entire GANDiffFace with \emph{identity preservation} filter $t_{ip} = 0.3$ (default) to evaluate the impact of the Diffusion model on intra-class variations. Then, we use the entire GANDiffFace framework with $t_{ip} = 0.2$ and $t_{ip} = 0.4$ to evaluate different intra-class variations.
Additionally, we consider subsets of two synthetic datasets, \emph{i.e.} SFace \cite{boutros2022sface} and DigiFace-1M \cite{bae2023digiface}. With the latter, we provide a comparison with a dataset based on 3D model.

\subsection{Real-world datasets}

We consider two real-world datasets widely used for face recognition, VGGFace2 \cite{Cao18} and IJB-C \cite{maze2018iarpa}. VGGFace2 is a large-scale dataset containing images from the web of around $9,000$ identities, with large variations in pose, age, illumination, ethnicity and profession. IJB-C contains around $3,000$ identities, with focus on occlusions and diversity of ethnicity and profession. According to IJB-C annotations, we remove multiple images taken from the same video and images with small faces. Both datasets have been discontinued, underlining the necessity of the generation of synthetic datasets with realistic intra-class variations.

For a fair comparison with our synthetic datasets, we filter out real images with a MagFace quality lower than $24.45$ \cite{meng2021magface}. This is the threshold used to eliminate the 10\% of images with the lowest magnitude during GANDiffFace identity generation (Section \ref{sec:3.1}). We are interested in the generation of datasets with high quality images. Datasets with low quality images can be obtained with data augmentation, and their evaluation is out of the scope of this work. 

\subsection{Similarity score distributions}
\begin{figure*}[t]
\begin{center}
\begin{subfigure}{0.25\textwidth}
    \centering
    \includegraphics[width=1\linewidth]{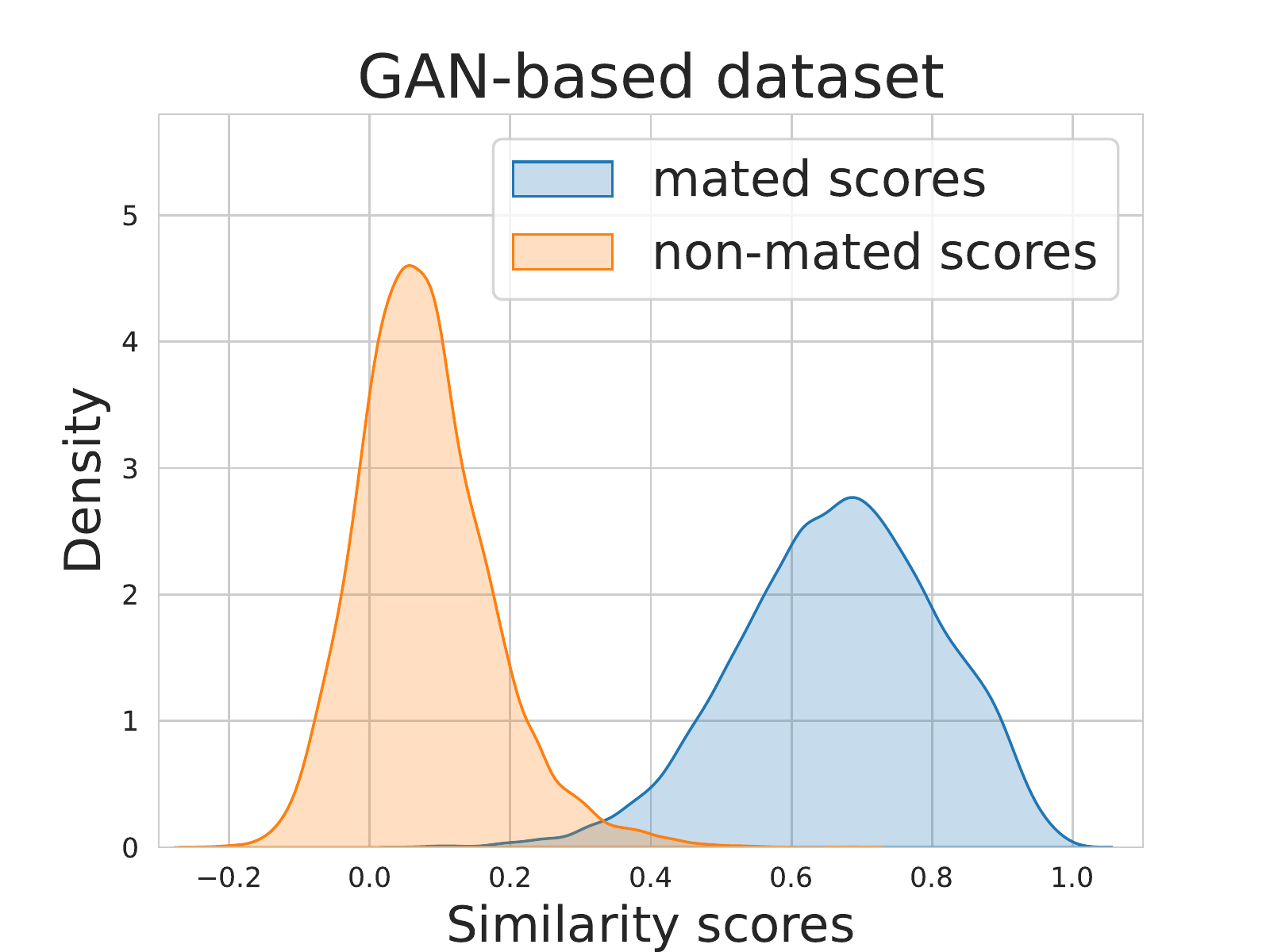}
    \label{fig:4a}
\end{subfigure}%
\begin{subfigure}{0.25\textwidth}
    \centering
    \includegraphics[width=1\linewidth]{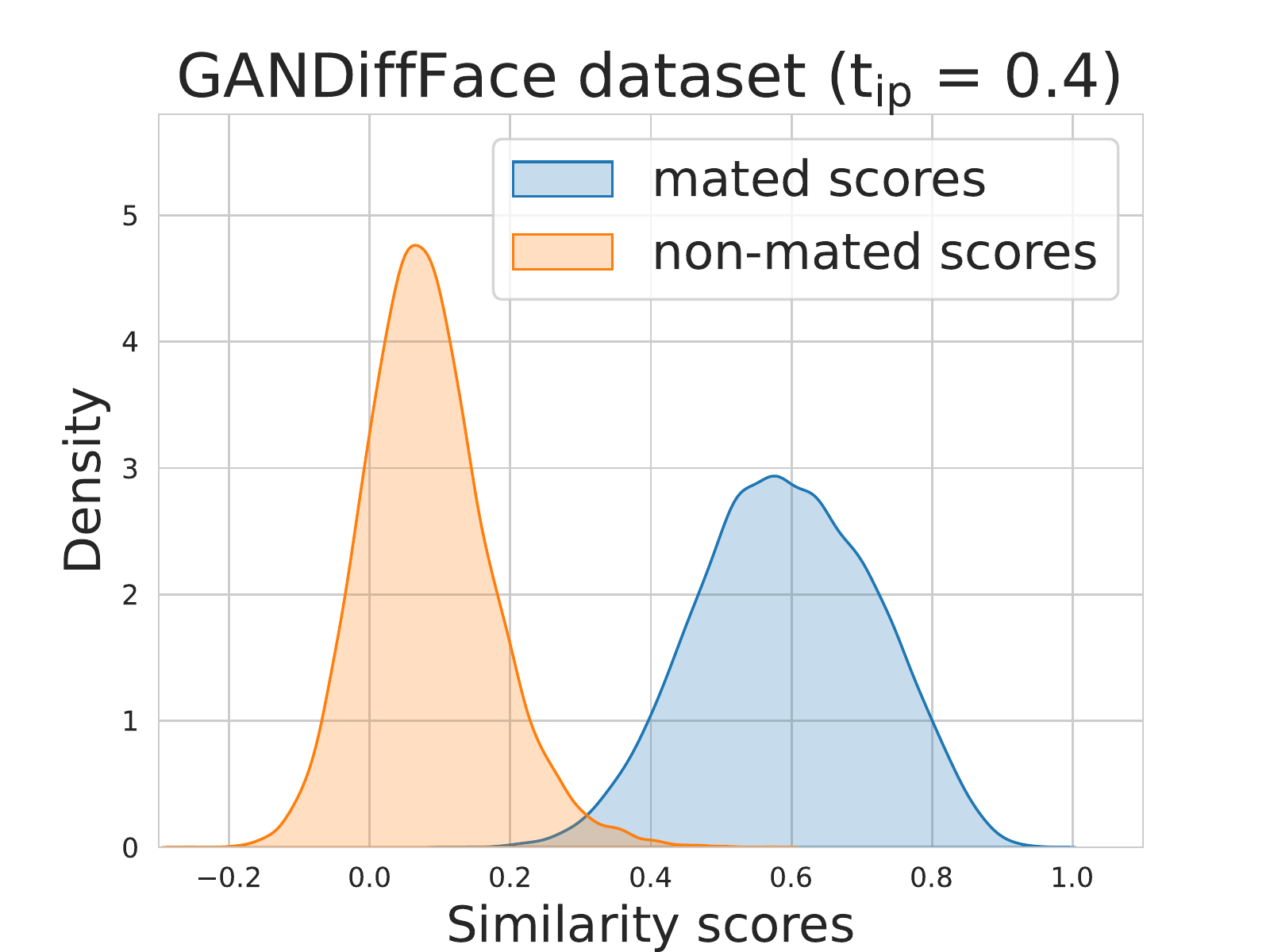}
    \label{fig:4b}
\end{subfigure}%
\begin{subfigure}{0.25\textwidth}
    \centering
    \includegraphics[width=1\linewidth]{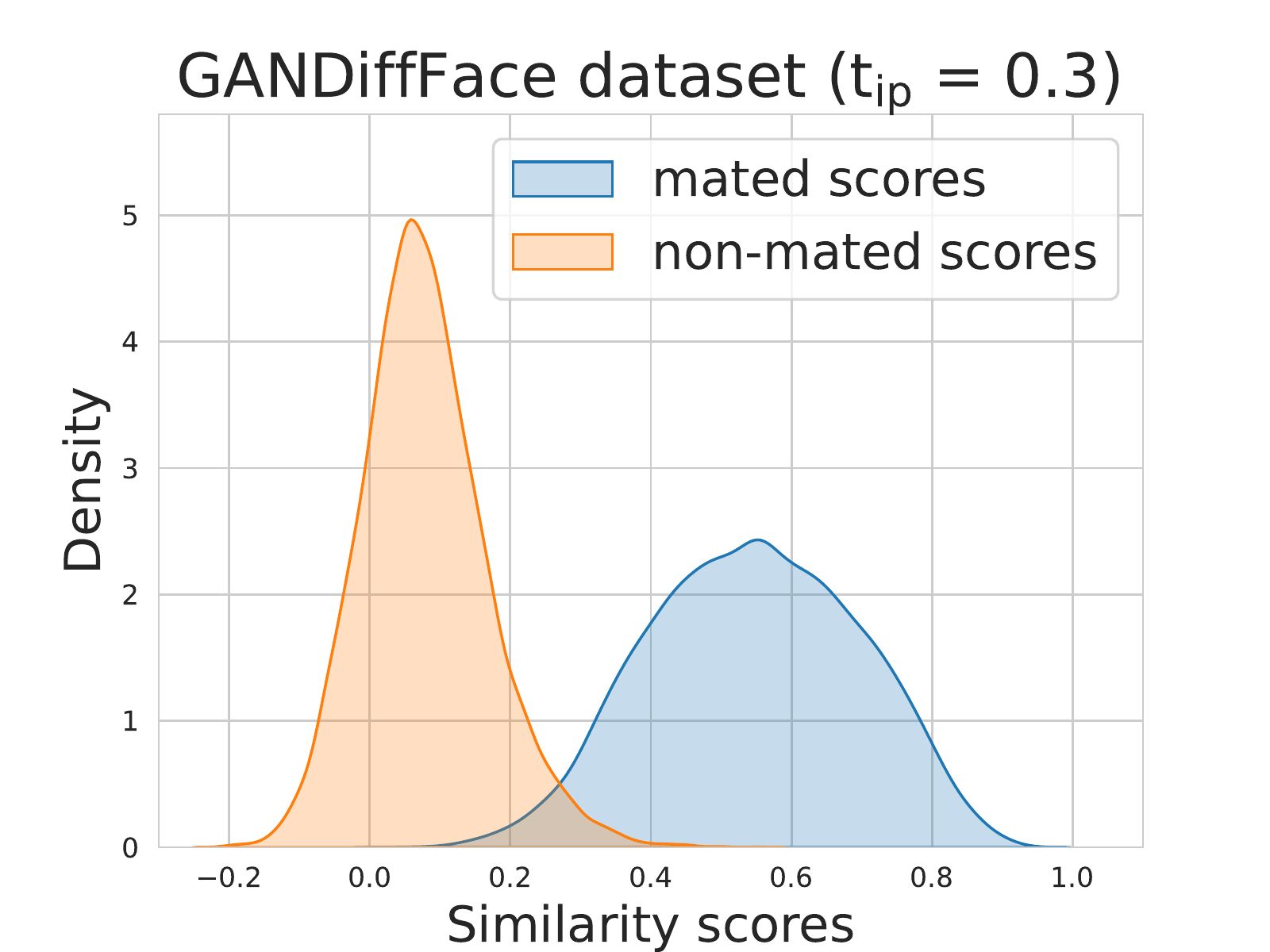}
    \label{fig:4c}
\end{subfigure}%
\begin{subfigure}{0.25\textwidth}
    \centering
    \includegraphics[width=1\linewidth]{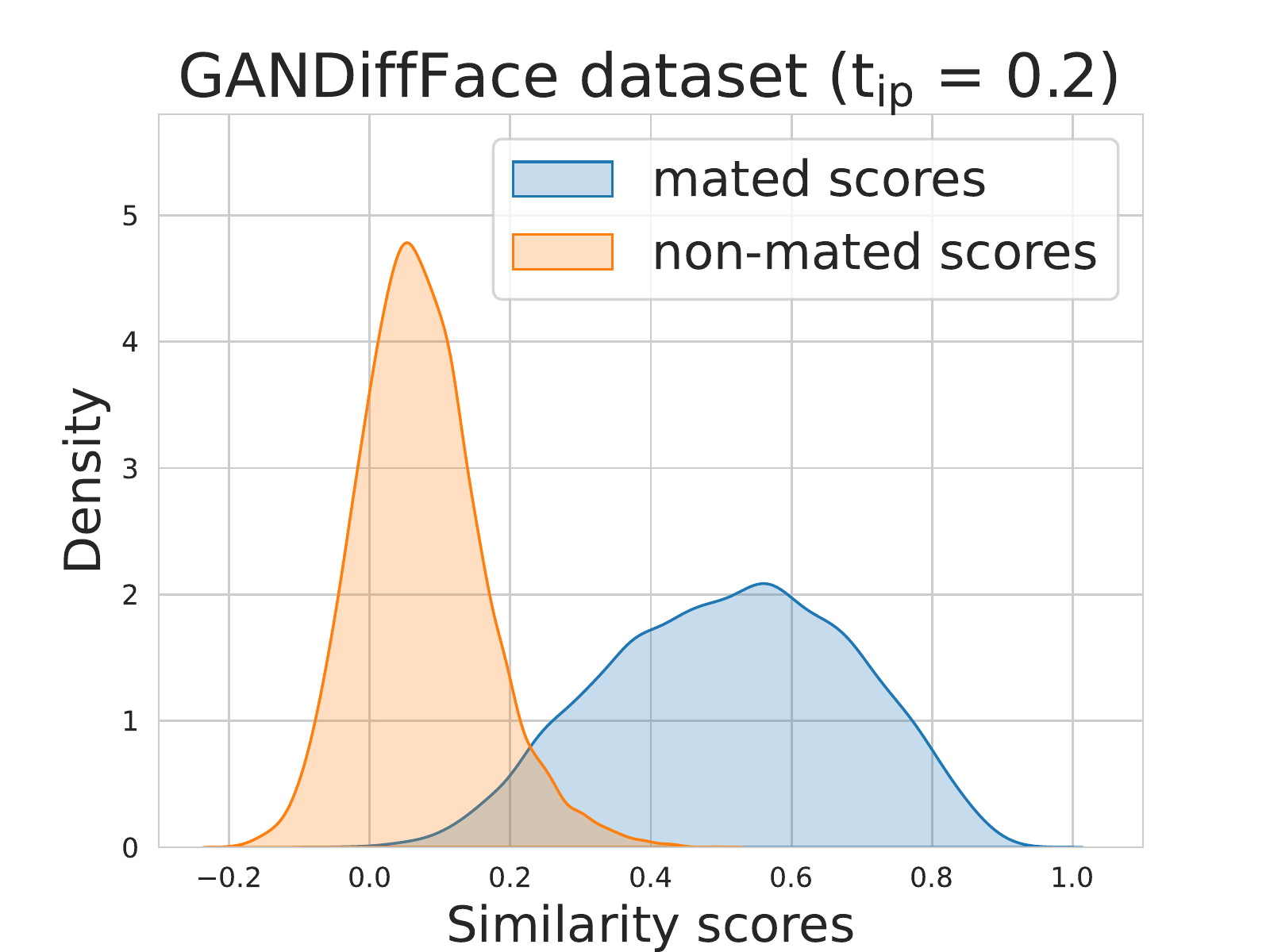}
    \label{fig:4d}
\end{subfigure}
\begin{subfigure}{0.25\textwidth}
    \centering
    \includegraphics[width=1\linewidth]{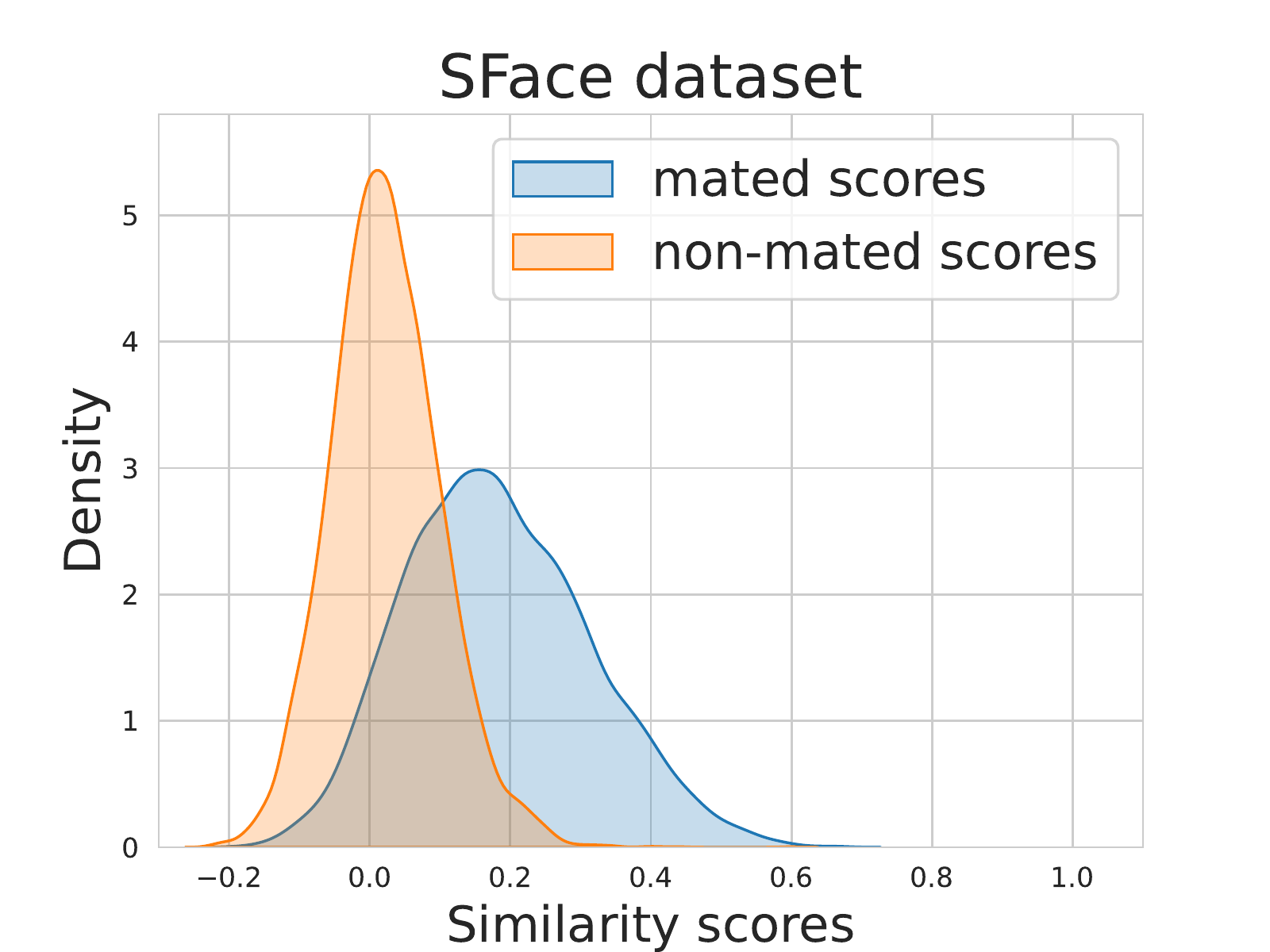}
    \label{fig:4e}
\end{subfigure}%
\begin{subfigure}{0.25\textwidth}
    \centering
    \includegraphics[width=1\linewidth]{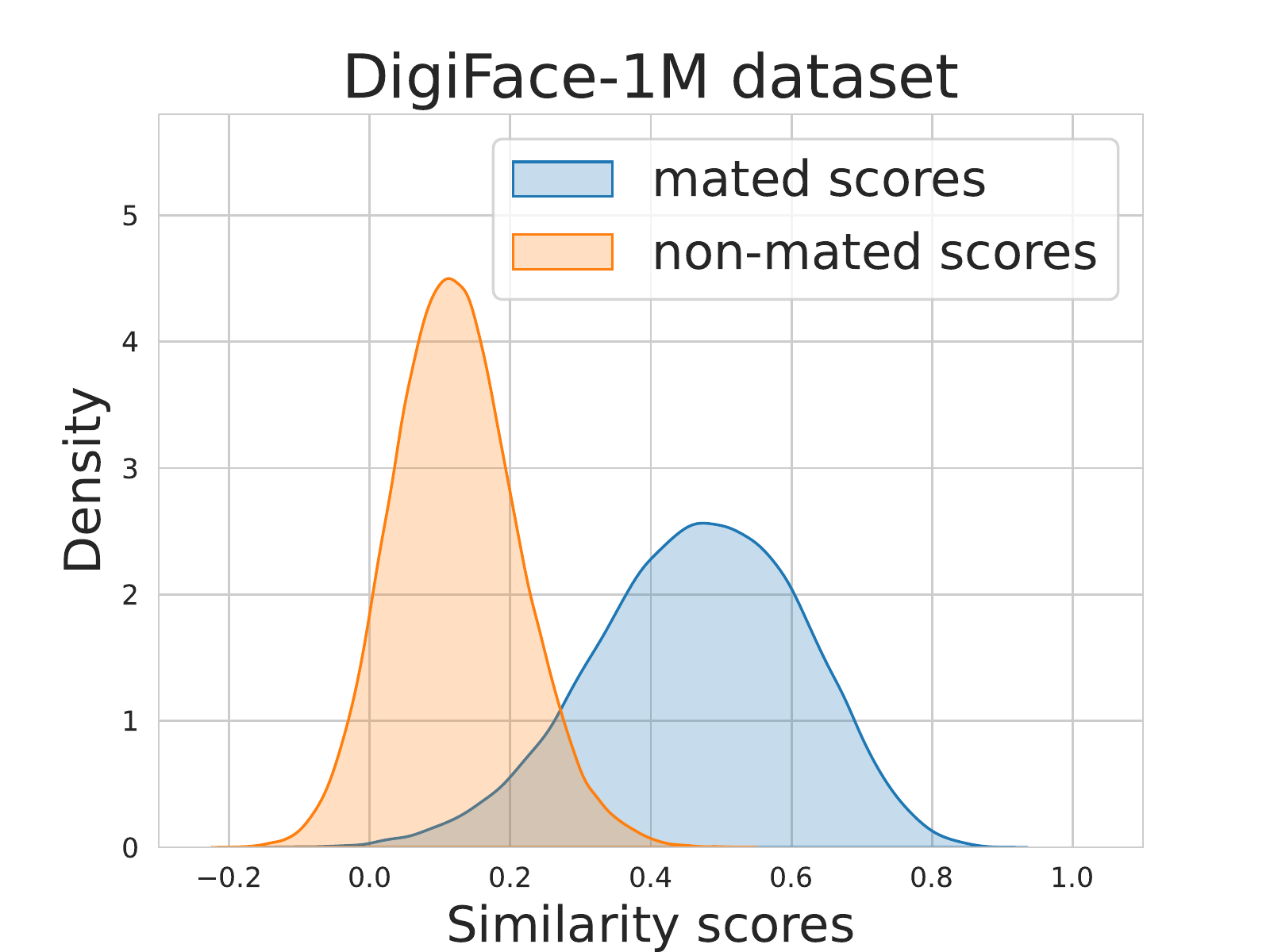}
    \label{fig:4f}
\end{subfigure}%
\begin{subfigure}{0.25\textwidth}
    \centering
    \includegraphics[width=1\linewidth]{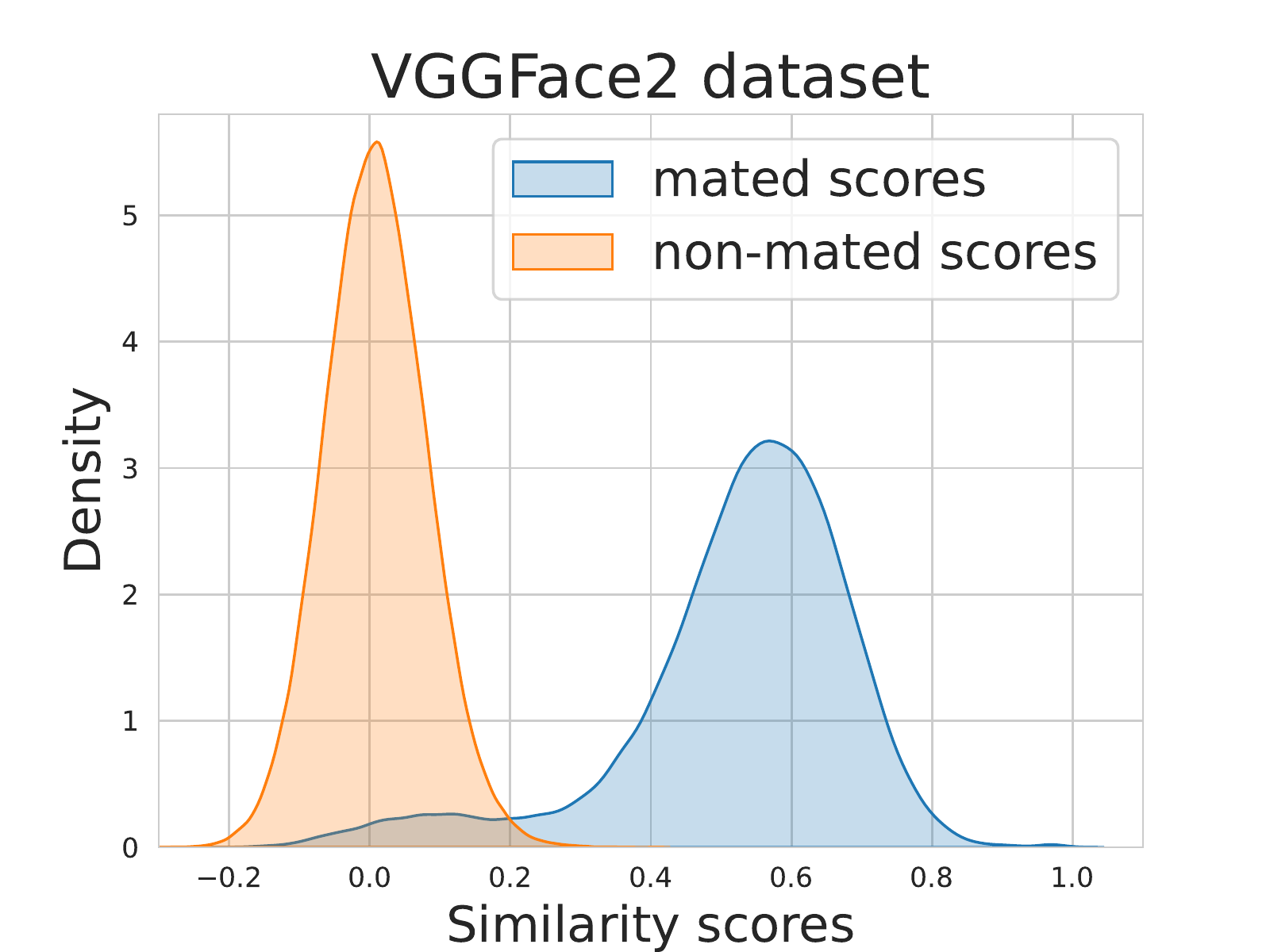}
    \label{fig:4g}
\end{subfigure}%
\begin{subfigure}{0.25\textwidth}
    \centering
    \includegraphics[width=1\linewidth]{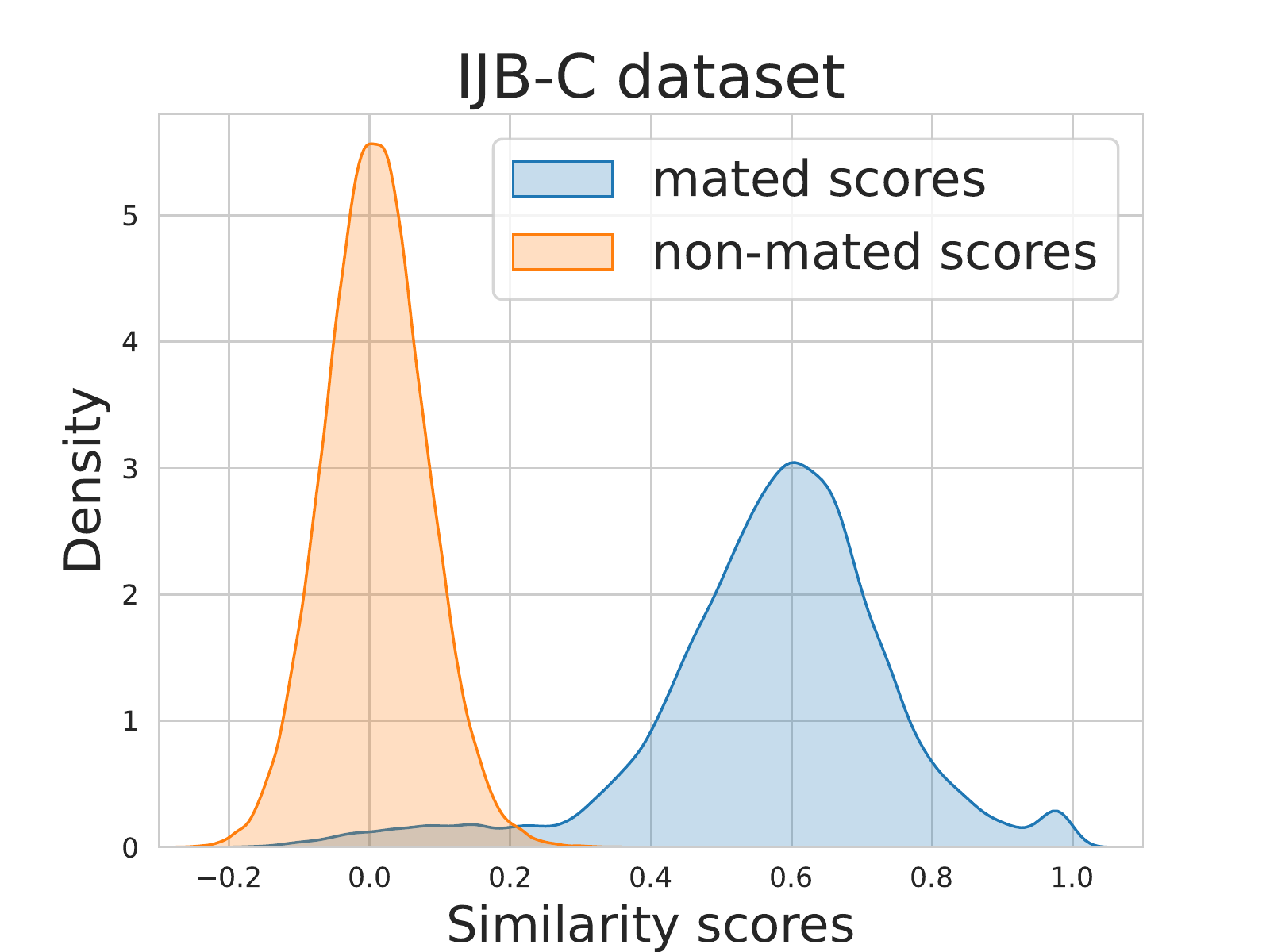}
    \label{fig:4h}
\end{subfigure}
\end{center}\vspace{-0.9cm}
\caption{Similarity score distributions obtained from mated and non-mated comparisons randomly selected from our synthetic datasets (first row), and other synthetic, SFace and DigiFace-1M, and real-world, VGGFace2 and IJB-C, datasets (second row).}\vspace{-0.4cm}
\label{fig:scores}
\end{figure*}

For each identity in real or synthetic datasets, we randomly select 10 images and generate $20$ mated and $20$ non-mated comparisons, and calculate the cosine similarity of their ArcFace (backbone iResNet100) embeddings \cite{deng2019arcface} (Figure \ref{fig:scores}). We use ArcFace as it is open source and widely used for face recognition. Then, we measure the diversity between synthetic and real score distributions, the latter as reference, with Kullback–Leibler (KL) divergence.

In Table \ref{tab:stats} we report the mean and standard deviation of mated and non-mated comparisons in the different datasets, as well as the number of identities. The reason behind the limited size of our GANDiffFace datasets is the high computational cost required for generation, but larger datasets can be produced. Analyzing the results, the use of a Diffusion model reduces the mean of mated scores from 0.67 (obtained with GAN-based module only) to 0.51, for $t_{ip} = 0.2$. This value is closer to the means of real datasets, \emph{i.e.} 0.52 for VGGFace2 and 0.57 for IJB-C. The IJB-C mean is affected by a peak in the score distribution for values close to 1, due to the comparison of images taken from the same video and not detected in the annotation file. According to Table \ref{tab:kl}, GANDiffFace with $t_{ip} \geq 0.3$ reproduces mated distributions similar to the real ones ($KL = 0.16$ from VGGFace2). We observe that the mated distribution of the GAN-based dataset is very far from the VGGFace2 one ($KL = 0.69$), while the mated comparisons with high score in IJB-C help to reduce KL divergence to 0.28 for GAN-based and 0.09 for GANDiffFace with $t_{ip} = 0.4$.

For non-mated comparisons, we observe that synthetic datasets present distributions skewed towards positive values, differently from real datasets (Figure \ref{fig:scores}). KL divergence is generally bigger for synthetic non-mated distributions, showing difficulty to reproduce realistic inter-class variations (Table \ref{tab:kl}). While no significant difference can be observed between the non-mated distributions of GAN-based and GANDiffFace ($t_{ip} = 0.3$) datasets, KL divergences are slightly higher for $t_{ip} = 0.4$, and decrease with $t_{ip} = 0.2$: from 0.48 to 0.42 with regard to VGGFace2, and from 0.43 to 0.37 with regard to IJB-C. This may be due to the inclusion in the synthetic dataset of images less similar to the GAN-based ones, which showed a slightly positive mean for non-mated comparison scores.

Finally, we observe worse score distributions for the synthetic SFace and DigiFace-1M datasets compared to the ones provided by GANDiffFace, with Equal Error Rates (EERs) about twice the real ones. This is reflected in higher KL divergences for mated (in case of SFace) and non-mated (in case of DigiFace-1M) score distributions, showing that these datasets fail to reproduce realistic intra and inter-class variations. However, SFace provides the best non-mated score distribution for synthetic datasets, with $KL = 0.18$ from VGGFace2 and $KL = 0.11$ from IJB-C.

\begin{table}
\begin{center}
\begin{footnotesize}
\setlength{\tabcolsep}{3pt}
\begin{tabular}{|c|c|c|c|c|}
\hline
\textbf{Dataset} & \textbf{Type} & \textbf{Id.} &\textbf{Mated scores}                & \textbf{Non-mated scores}           \\ \hline \hline

GAN-based & Syn & 700 & $0.67 \pm 0.14$ & $0.08 \pm 0.10$\\
\begin{tabular}[c]{@{}c@{}}GANDiffFace\\($t_{ip} = 0.4$)\end{tabular} & Syn & 700 & $0.59 \pm 0.12$ & $0.08 \pm 0.09$ \\
\begin{tabular}[c]{@{}c@{}}GANDiffFace\\($t_{ip} = 0.3$)\end{tabular} & Syn & 700 & $0.55 \pm 0.15$ & $0.08 \pm 0.09$ \\
\begin{tabular}[c]{@{}c@{}}GANDiffFace\\($t_{ip} = 0.2$)\end{tabular} & Syn & 700 & $0.51 \pm 0.17$ & $0.07 \pm 0.09$ \\
SFace & Syn & 411 & $0.18 \pm 0.13$ & $0.02 \pm 0.08$ \\
DigiFace-1M & Syn & 2,000 & $0.47 \pm 0.15$ & $0.12 \pm 0.09$ \\ \hline
VGGFace2 & Real & 8,515 & $0.52 \pm 0.16$ & $0.01 \pm 0.07$ \\
IJB-C & Real & 2,557 & $0.57 \pm 0.17$ & $0.01 \pm 0.07$ \\ \hline
\end{tabular}
\end{footnotesize}
\end{center}\vspace{-0.4cm}
\caption{Number of identities and means of mated/non-mated score distributions of synthetic and real datasets.}
\label{tab:stats}\vspace{-0.4cm}
\end{table}

\begin{table}
\begin{center}
\begin{footnotesize}
\setlength{\tabcolsep}{4.4pt}
\begin{tabular}{|c|cc|cc|c|}
\hline
\multirow{2}{*}{\textbf{Dataset}} & \multicolumn{2}{c|}{\textbf{Mated scores}}                 & \multicolumn{2}{c|}{\textbf{Non-mated scores}} & \multirow{2}{*}{\textbf{EER}}            \\
                              & \textbf{VGG2}        & \textbf{IJB-C} & \textbf{VGG2}        & \textbf{IJB-C} & \\ \hline \hline

GAN-based & 0.69 & 0.28 & 0.48 & 0.42 & 1.49\%\\
\begin{tabular}[c]{@{}c@{}}GANDiffFace\\($t_{ip} = 0.4$)\end{tabular} & \textbf{0.16} & \textbf{0.09} & 0.52 & 0.46 & 1.25\% \\
\begin{tabular}[c]{@{}c@{}}GANDiffFace\\($t_{ip} = 0.3$)\end{tabular} & \textbf{0.16} & 0.16 & 0.48 & 0.43 & 2.74\% \\
\begin{tabular}[c]{@{}c@{}}GANDiffFace\\($t_{ip} = 0.2$)\end{tabular} & 0.23 & 0.28 & 0.42 & 0.37 & 5.11\% \\
SFace & 1.72 & 2.13 & \textbf{0.18} & \textbf{0.11} & 22.53\%\\
DigiFace-1M & 0.21 & 0.41 & 1.05 & 1.02 & 7.92\%\\ \hline
VGGFace2 & - & 0.11 & - & 0.01 & 4.51\%\\
IJB-C & 0.15 & - & 0.01 & - & 3.22\% \\ \hline
\end{tabular}
\end{footnotesize}
\end{center}\vspace{-0.4cm}
\caption{KL divergences of the distributions of each dataset from the real ones provided by VGGFace2 and IJB-C. }
\label{tab:kl}\vspace{-0.4cm}
\end{table}

\section{Conclusion} \label{sec:5}
This study has proposed GANDiffFace, a novel framework that combines the advantages of GAN and Diffusion models to generate synthetic datasets for face recognition with some desired properties. The use of a GAN model for identity generation, \emph{i.e.} StyleGAN3, allows to synthesize images of human faces with high realism, and manipulate the latent space to provide a fair representation of $70$ demographic groups. The addition of a Diffusion model, \emph{i.e.} Stable Diffusion, personalized for specific identities with DreamBooth, allows the fully synthetic generation of a dataset with realistic intra-class variations.

A limitation of GANDiffFace consists in the high computational cost required to fine-tune identity-specific Diffusion models. This was the main reason for the generation of $700$ identities, but many more can be generated. Also, Diffusion images present some artifacts observable at human level. Nevertheless, they usually affect parts of human bodies such as hands that are cut out for face recognition. 

In future works, we plan to use the synthetic dataset generated with GANDiffFace to deploy face recognition systems, given its desired properties of realistic intra-class variations and fair representation of multiple demographic groups. Also, future works can focus on reducing the KL divergence from non-mated score distributions of real datasets, to reproduce more accurately real-world inter-class variations.

\newpage

\section*{Acknowledgments}
This project has received funding from the European Union’s Horizon 2020 research and innovation programme under the Marie Sk\l{}odowska-Curie grant agreement No 860813 - TReSPAsS-ETN. This study is supported by the project INTER-ACTION (PID2021- 126521OB-I00 MICINN/FEDER). This research is based upon work supported by the Hessian Ministry of the Interior and Sport in the course of the Bio4ensics project and by the German Federal Ministry of Education and Research and the Hessian Ministry of Higher Education, Research, Science and the Arts within their joint support of the National Research Center for Applied Cybersecurity ATHENE.

{\small
\bibliographystyle{ieee_fullname}
\bibliography{manuscript}
}

\end{document}